\documentclass[10pt,twocolumn,letterpaper]{article}

\usepackage{overpic}
\usepackage{float}
\usepackage{bm}
\usepackage{multicol}
\usepackage{multirow}
\usepackage{pifont}
\usepackage{color}
\usepackage{overpic}
\usepackage{arydshln}
\usepackage{makecell}
\usepackage{times}
\usepackage{epsfig}
\usepackage[T1]{fontenc}
\usepackage{hhline}
\usepackage{pifont}
\usepackage[shortlabels,inline]{enumitem}
\usepackage{colortbl}
\usepackage{verbatim}
\usepackage{bm}
\usepackage{fancyhdr}
\usepackage{multirow}
\usepackage[pagenumbers]{iccv} 

\definecolor{iccvblue}{rgb}{0.21,0.49,0.74}
\usepackage[pagebackref,breaklinks,colorlinks,allcolors=iccvblue]{hyperref}

\title{SIGMAN: \underline{S}caling 3D Human \underline{G}aussian Generation with \underline{M}illions of \underline{A}ssets}

\author{Yuhang Yang$^{1,2,*}$, Fengqi Liu$^{2,3*}$, Yixing Lu$^{4}$, Qin Zhao$^{2}$, Pingyu Wu$^{1}$, Wei Zhai$^{1,\dagger}$, Ran Yi$^{3}$ \\ Yang Cao$^{1}$, Lizhuang Ma$^{3}$, Zheng-Jun Zha$^{1}$, Junting Dong$^{2,\dagger}$\\
{$^{1}$~USTC \textit{} $^{2}$~Shanghai AI Lab \textit{} $^{3}$~SJTU \textit{} $^{4}$~CMU}\\
$*$Equal Contribution \textit{} $\dagger$Corresponding Author\\
\href{https://yyvhang.github.io/SIGMAN_3D/}{https://yyvhang.github.io/SIGMAN\_3D/}
}
\begin{document}

\twocolumn[{%
         \renewcommand\twocolumn[1][]{#1}%
         \maketitle
         \begin{center}
            \centering
            \includegraphics[width=0.97\textwidth]{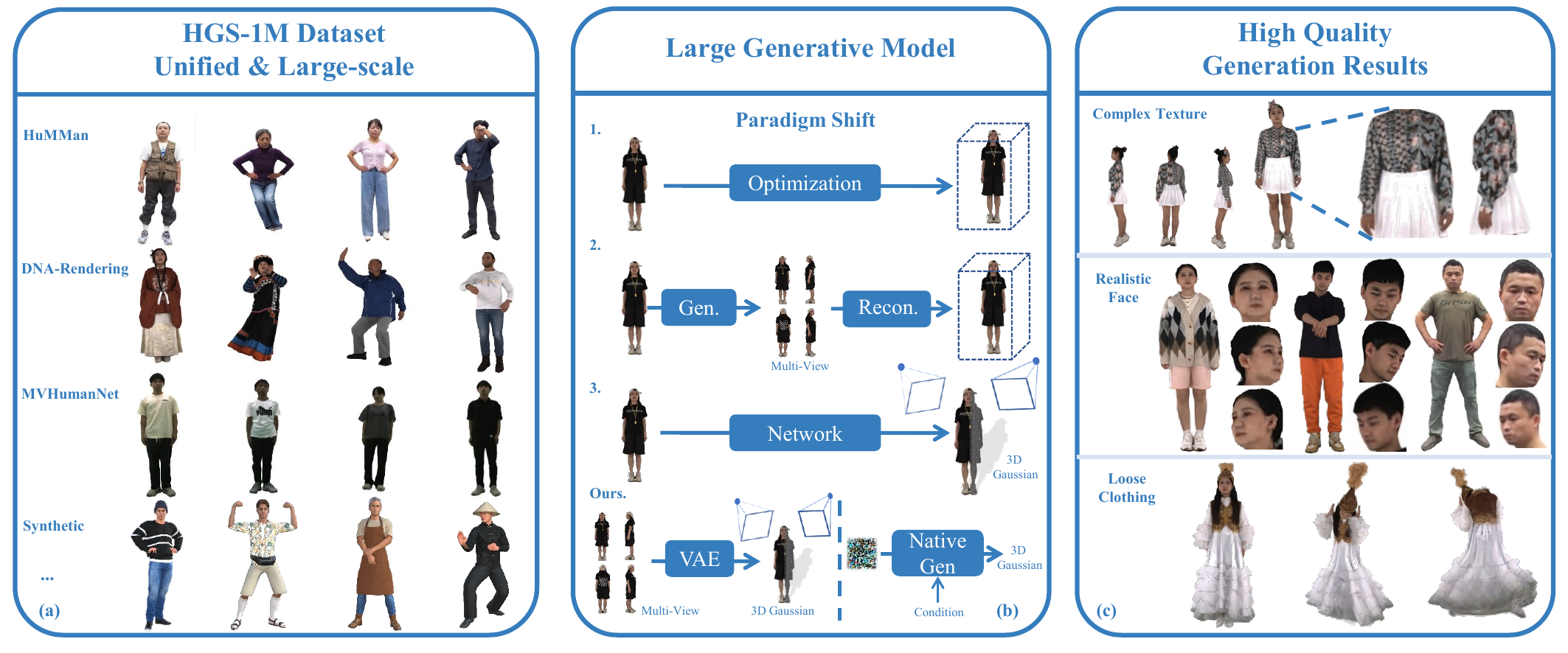}
            \captionof{figure}{\textbf{(a)}. In this work, we construct a large-scale, unified 3D Human Gaussian Dataset, called HGS-1M, to support \textbf{(b)} the large-scale generation model for 3D human Gaussian generation. \textbf{(c)} This paradigm, with large-scale data, produces high-quality 3D human Gaussians that exhibit complex textures, facial details, and realistic deformation of loose clothing.}
            \label{fig:teaser}
         \end{center}
}]

\begin{abstract}
3D human digitization has long been a highly pursued yet challenging task. Existing methods aim to generate high-quality 3D digital humans from single or multiple views, but remain primarily constrained by current paradigms and the scarcity of 3D human assets. Specifically, recent approaches fall into several paradigms: optimization-based and feed-forward (both single-view regression and multi-view generation with reconstruction). However, they are limited by slow speed, low quality, cascade reasoning, and ambiguity in mapping low-dimensional planes to high-dimensional space due to occlusion and invisibility, respectively. Furthermore, existing 3D human assets remain small-scale, insufficient for large-scale training. To address these challenges, we propose a latent space generation paradigm for 3D human digitization, which involves compressing multi-view images into Gaussians via a UV-structured VAE, along with DiT-based conditional generation, we transform the ill-posed low-to-high-dimensional mapping problem into a learnable distribution shift, which also supports end-to-end inference. In addition, we employ the multi-view optimization approach combined with synthetic data to construct the HGS-1M dataset, which contains $1$ million 3D Gaussian assets to support the large-scale training. Experimental results demonstrate that our paradigm, powered by large-scale training, produces high-quality 3D human Gaussians with intricate textures, facial details, and loose clothing deformation.
\end{abstract}
\section{Introduction}
\label{sec:intro}
Human digitization has always been a central topic in the generation and reconstruction field, it supports various applications such as gaming, AR/VR, etc. For 3D human digitization, the task usually involves mapping the input (\eg, single image, multi images, text) to 3D human assets, which then could be animated through rigging. Compared to pixel-level character digitalization \cite{hu2023animateanyone,men2024mimo,esser2024scaling,chen2025janus,agarwal2025cosmos,kong2024hunyuanvideo,genmo2024mochi,li2024dispose}, 3D character digitalization offers advantages such as multi-view consistency and more stable controllability.

Recently, studies on 3D human digitization primarily focus on image-to-3D human, and these methods can be summarized into the following paradigms. \textbf{1)} The optimization-based approach \cite{xiu2022icon,xiu2023econ,zhang2023global,zhang2024sifu}: which typically utilizes geometric constraints and human priors to optimize each character, the limitations of this type of method are slow speed and limited quality; \textbf{2)} Feed-forward: this includes two manners, one of which is the two-stage cascade manner \cite{li2024pshuman,weng2024template,wang2025wonderhuman}, this kind of method first generates multiple views that correspond to the input view, and then takes these views to reconstruct the 3D human. Its cascade structure requires both stages to work well, however, humans possess high diversity, generating multiple views from a single perspective while maintaining consistency is highly unstable, affecting the final quality. The other manner \cite{zhuang2024idol,kwon2024generalizable,chen2024generalizable,kwon2021neural,qiu2025LHM,hu2024gaussianavatar} involves taking a single view or sparse views as the input and employing a network to directly predict the 3D human (represented in 3D Gaussian Splatting \cite{kerbl20233d}). Although these methods are faster and further improve the quality of digitalization, it requires the network to learn the mapping from 2D-pixel planes to 3D geometric space through limited views, resulting in occlusion and invisibility problems, which makes it hard for the network to learn the complete mapping and low quality shows up on non-visible sides.

To address the above limitations, in this paper, we use 3D Gaussian Splatting as the representation, pushing 3D human digitization towards the paradigm of 3D native generation in latent space. Specifically, our paradigm involves training a VAE model \cite{pu2016variational} to compress multi-view consistent images into latent representations and then decode them into Gaussian attributes. Afterward, we train a conditional DiT-based \cite{peebles2023scalable} generation model, which takes a single image or text as the condition to generate the 3D human Gaussian. In this paradigm, the VAE stage has multiple consistent and comprehensive views as the inputs, providing sufficient geometric contexts and enabling the Gaussian latent to represent high-quality geometry, appearance details, and pose-dependent deformation. Turning to the generation, the learning objective shifts to capture the transition from conditional features to latents, it avoids learning the hard mapping from low-dimension (2D plane) to high-dimension (3D space). Moreover, in the inference stage, only a certain number of inference steps are needed to generate the corresponding latent, refrain from the cascade structure, and does not need per-case optimization.

The pose and appearance of humans have strong diversity, and there are many high-frequency details, \eg, complex textures of clothes and faces. Directly compressing multi-view images to 3D human Gaussians is non-trivial. To achieve this, we introduce a UV-structured latent representation, which takes parametric model SMPL-X \cite{SMPL-X:2019} as human priors and leverages multi-view geometric contexts as references, employing learnable tokens defined in the SMPL-X UV coordinates to query from geometric contexts, thus modeling the latent, it can then be decoded to Gaussian attributes on the UV coordinate, by back-projecting onto the SMPLX template, we obtain the canonical human Gaussian, and the posed 3D human Gaussian could be obtained through LBS. This manner avoids directly learning the explicit geometric structure, which is conducive to the learning of appearance details. In addition, the LBS-driven method enables the latent in the VAE stage to represent the corresponding Gaussian properties under different poses and capture the implicit pose-related deformation. Furthermore, we also introduce UV-initialization and Conv-Atten dual branch mechanisms in the encoding stage to accelerate convergence and enhance the detail learning. In the generation stage, we adopt the MM-DiT architecture and a human-specific encoder \cite{khirodkar2024sapiens} to better learn the transformation from conditional features to Gaussian latents.

In addition to the learning paradigm, a particularly challenging issue is that 3D human assets are very scarce, which can not support large-scale training. However, the multi-view loop optimization method can reconstruct high-quality 3D human Gaussians from dense views. In light of this, we use AnimatableGaussians \cite{li2024animatablegaussians} to build a 3D human Gaussian dataset from multi-view human datasets \cite{yu2021function4d,cheng2023dna,cai2022humman,zheng2019deephuman,xiong2024mvhumannet,han2023high}, to expand the data diversity, we also combine synthetic data. Eventually, we construct the HGS-1M dataset, which contains $1$ million 3D human Gaussians, including diverse races, genders, age groups, appearances, clothing, and postures. This large-scale dataset could support the training of large-scale models and serve as the new test bed for 3D human digitization.

\par The contributions are summarized as follows:
\begin{itemize}[leftmargin=15pt,topsep=0pt,itemsep=2pt]
    \item[\textbf{1)}] We introduce a paradigm of using large-scale latent generation for 3D human digitization. Through two-stage learning of compression and generation, we transform the challenging low-to-high-dimensional mapping problem into a conditional feature-to-latent transfer task, enabling high-quality modeling of 3D humans.
    \item[\textbf{2)}] We propose the UV-structured latent representation, which leverages the parametric SMPL-X as priors to enhance VAE’s learning of appearance details and pose-related deformations. With the MMDiT-based generation, we perform native 3D human generation.
    \item[\textbf{3)}] We construct the HGS-1M dataset, which contains $1$ million 3D human Gaussians, encompassing diverse races, genders, ages, appearances, clothing, and postures. It supports large-scale training and serves as the new test bed for 3D human digitization.
    \item[\textbf{4)}] Experiments show that large-scale data and model improve the quality of 3D human digitization, highlighting the superiority over previous methods and paradigms.
\end{itemize}

\section{Related Work}
\begin{figure}
    \centering
    \includegraphics[width=\linewidth]{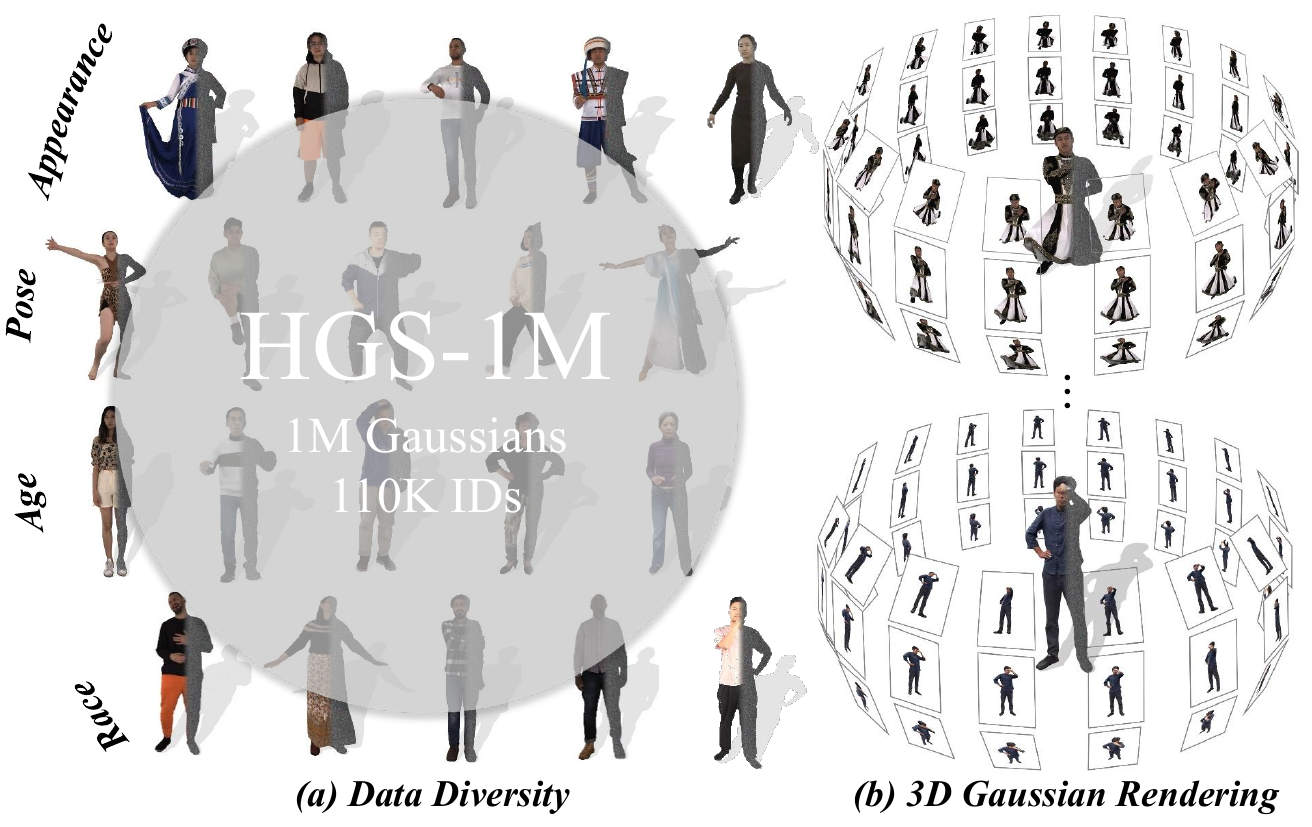}
    \caption{\textbf{HGM-1M Dataset}. The constructed HGS-1M dataset, it contains $1$ million 3D Gaussian human assets of different ages, races, appearances, and poses. It supports free-view rendering.}
    \label{fig:dataset}
\end{figure}

\subsection{3D Human Digitization}
For 3D human creation, a branch adopts implicit representation \cite{saito2019pifu,saito2020pifuhd,zheng2021pamir,hu2023sherf,huang2023one,kwon2021neural,kwon2023neural,peng2021animatable,lu2025gas,zhao2022humannerf} and explicit human prior for optimization \cite{dong2022totalselfscan,zhang2024sifu,zhang2023global}, calculating each instance individually. While these methods enhance the robustness to a certain extent, they take relatively longer to get results for each input, lack generalizability, and deliver limited quality. To enable rapid digitization and obtain high-quality, photo-realistic 3D humans, some approaches gradually shift to the feed-forward manner, including using a regression network \cite{zheng2024gps,zhuang2024idol,kwon2024generalizable, chen2024generalizable,kwon2023deliffas,qiu2025LHM} and cascaded generation with reconstruction \cite{li2024pshuman,weng2024template,tang2024lgm,peng2024charactergen,dong2023ivs}. For the regression-based method, part of it needs multiple views as the input, facing significant constraints during inference, whereas single-view approaches relax this limitation but necessitate the network to learn mappings from low-dimensional planes to high-dimensional space, encountering occlusion and invisibility issues. The cascade type that generates multi-view images at first and then reconstructs the 3D assets, it requires view-consistent generation in the first stage, which is unstable and remains challenging, affecting the subsequent reconstruction. Recent work E$^{3}$Gen \cite{zhang2024e3gen} explores native 3D generation for human digitization but focuses solely on unconditional generation. For the animation, some methods \cite{zhuang2024idol,zhang2024e3gen,qiu2025LHM} adopt the LBS manner, using SMPL(X)-based motion sequences \cite{xiao2025motionstreamer,zhang2023generating,tevet2022human} to drive, other method like DRIVE \cite{sun2024drive} employs rigging for the generated 3D human Gaussian, driving the character. In addition to obtaining 3D humans from images, a few works also explore generating 3D humans from text \cite{dong2024tela,liu2023humangaussian,kolotouros2023dreamhuman,gong2024text2avatar}. Nonetheless, prior research predominantly focuses on small-scale datasets, lacking large-scale unified datasets and benchmarks to advance the 3D human digitization. In this work, we define a new paradigm for obtaining 3D humans and construct a large-scale, unified human Gaussian dataset to promote the field forward.

\subsection{Large Model for Generation}
The LLM \cite{ouyang2022training} opens the large-scale generation era, and the visual field also gradually shifts to this paradigm, however, the number of visual tokens is huge, and the generation speed and computational complexity are greatly challenging. Stable diffusion \cite{podell2023sdxl} starts the latent generation route, which compresses the raw visual data to latent and then employs diffusion models to generate the latent. It has achieved promising results in fields like image generation \cite{kolors, liu2024playground, flux2024}, video generation \cite{kong2024hunyuanvideo, genmo2024mochi, agarwal2025cosmos}, and 3D generation \cite{zhang2024clay, hunyuan3d22025tencent, xiang2024structured}. For this paradigm, there are several key aspects: large-scale and high-quality data; an appropriate training objective to learn the transfer between conditional features and latents; and a large-scale model to learn the target distribution and ensure the generated quality.

\section{HGS-1M Dataset}
\begin{table}[t]
  \centering
  \resizebox{0.47\textwidth}{!}{
\begin{tabular}{lrrcc}
\toprule
\multicolumn{1}{l}{\textbf{Dataset}} & \multicolumn{1}{r}{\textbf{\#Frames}} & \multicolumn{1}{r}{\textbf{\#ID}} & \multicolumn{1}{c}{\textbf{\# View}}  & \multicolumn{1}{c}{\textbf{\# 3D Asset}} \\

\midrule
      Human3.6M \cite{ionescu2013human3}&   3.6M     & 11   &4     & -   \\
      HUMBI \cite{yu2020humbi}&  26M     & 772     &107   & -  \\
      HuMMan \cite{cai2022humman}& 60M   &  1000    & 10   & - \\
      ActorsHQ \cite{icsik2023humanrf}&  40K   &  8    & 160  & - \\
      DNA-Rendering \cite{cheng2023dna}&  67.5M     &  500    & 60  & -  \\
      MVHumanNet \cite{xiong2024mvhumannet}&   \textbf{645.1M}    &  4500    & 48  & -  \\
      HuGe100K \cite{zhuang2024idol}  & 2.4M & 100K & 24 & - \\
\midrule
    THuman2.1 \cite{yu2021function4d}  & -   & 2500    & \textbf{Free} & 2500  \\
    2K2K \cite{han2023high}   & - & 2050 & \textbf{Free}   & 2050 \\
    \midrule
       \textbf{HGS-1M} & $>$ 90M & \textbf{110K}  & \textbf{Free} & \textbf{1M}\\
    \bottomrule
    
    \end{tabular}
     }
    \caption{Comparison of various attributes of multiple datasets. HGS1-M contains the most IDs and large-scale 3D assets that support rendering from any perspective.}
\label{tab:datasets}
\end{table}

Currently, the public 3D whole human assets are very limited \cite{zheng2019deephuman,yu2021function4d, han2023high}, which cannot meet the requirements for large-scale training. Although multi-view human datasets \cite{xiong2024mvhumannet,cheng2023dna,cai2022humman} can provide a large amount of human data, the quantity and pose of the camera are different in each dataset, and the views of some datasets are not dense enough to serve as supervision. To train a unified Gaussian generative model, unifying these data and getting as many perspectives as possible is necessary. Fortunately, the multi-view optimization method can reconstruct high-quality Gaussian data from multi-view human datasets, thus, we employ AnimatableGaussians \cite{li2024animatablegaussians} to merge multiple multi-view human datasets and reconstruct sequences of human Gaussians, each sequence taking around $20$ 4090 GPU hours for optimization. While the total computational overhead is huge, this approach allows us to obtain high-quality human Gaussian data and could render it at free-view for supervision. Next, we place each human Gaussian at the origin and unify the Gaussian position based on the corresponding SMPL-X root rotation and translation. For each human Gaussian, we select 90 views for rendering, including $30$ horizontal, $30$ upward, and $30$ downward perspectives, shown in Fig. \ref{fig:dataset} For the small-scale data \eg, Thuman2.1 \cite{yu2021function4d} and 2K2K \cite{han2023high}, we also move the asset to the origin and render it using the same cameras. Besides, we collect over $100k$ synthetic data and convert them into the same representation to further expand the diversity. Eventually, we construct the dataset HGS-1M, which contains $1$ million 3D human Gaussians with diverse races, genders, age groups, appearances, clothing, and postures, shown in Fig. ~\ref{fig:dataset}. Compared with previous datasets, it contains the largest amount of 3D human assets and has rich IDs and poses (Tab. \ref{tab:datasets}). Moreover, it can be rendered from free-view, which supports large-scale training and evaluation.
\begin{figure*}[t]
	\centering
        \scriptsize
	\begin{overpic}[width=1.\linewidth]{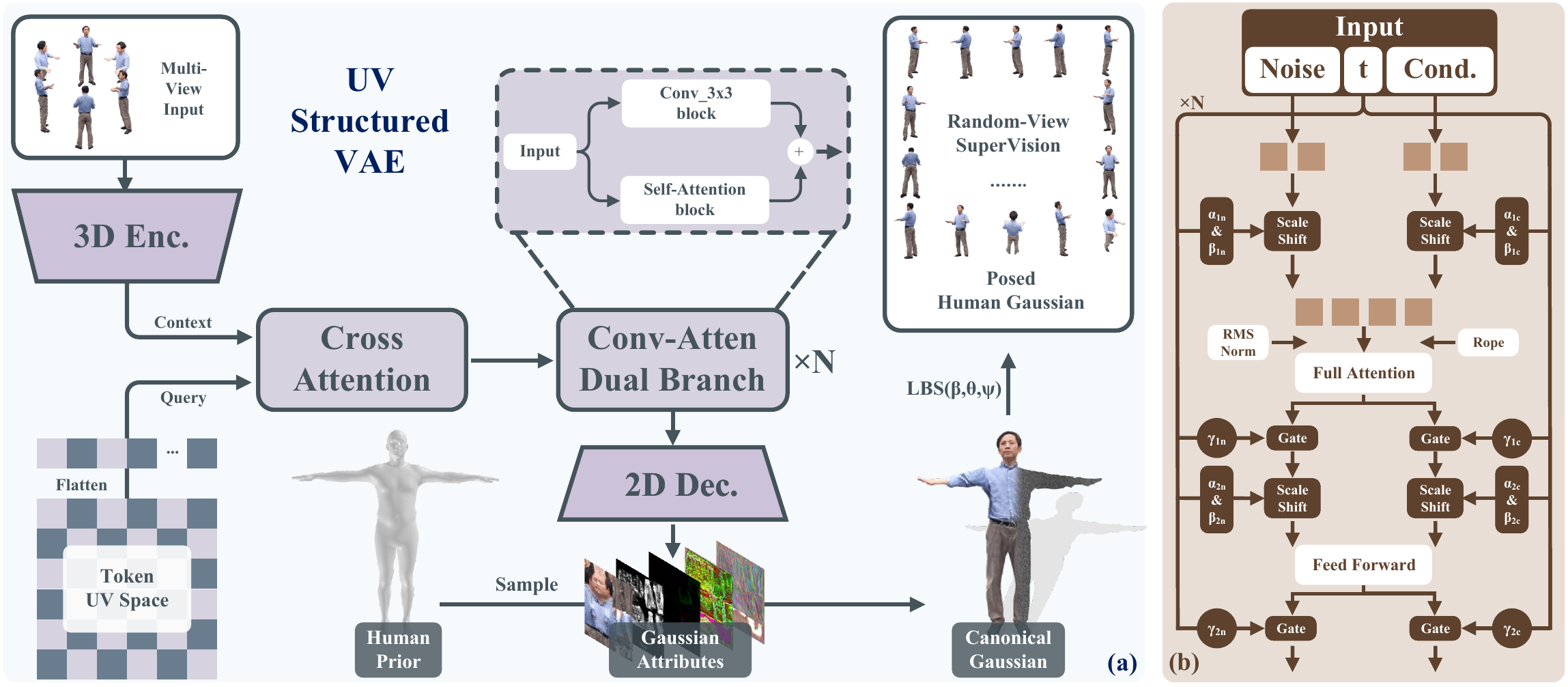}
 
	\end{overpic}
	\caption{\textbf{Method.} The pipeline of our method. \textbf{(a)}. The UV-structured VAE, which uses human priors to define learnable tokens in UV space and takes them to query multi-view contexts to model the Gaussian latent, then, the latent is decoded into human Gaussians in canonical space and could be driven by differentiable LBS to obtain the final posed human Gaussian. \textbf{(b)}. The MM-DiT architecture, it treats the conditional sequence and noise as a whole sequence to complete controllable 3D human Gaussian generation.}
 \label{fig:method}
\end{figure*}

\section{Method}
\label{Sec:method}
Our goal is to generate 3D human Gaussians from the instruction (single image or text), the method consists of a UV-structured VAE, and an MMDiT-based generation model, shown in Fig. \ref{fig:method}. The VAE compresses multi-view images into latent Gaussian attributes in the UV structural space, and the DiT aims to capture the transformation from conditional features to the compressed Gaussian latents. Through this way, the challenging mapping from sparse 2D image planes to 3D Gaussian space is thereby transformed into a distribution shift process.

\subsection{Preliminary}
\label{Sec:preliminary}
\textbf{SMPL-X} \cite{SMPL-X:2019} is a parametric human model that represents human body with a deformable mesh, formulated as:
\begin{equation}
\label{Eq.1}
\left.M(\beta, \theta, \psi)=\operatorname{LBS}\left(T_P(\beta, \theta, \psi)\right), J(\beta), \theta, \mathcal{W}\right),
\end{equation}
where $\beta, \theta, \phi$ represent shape, pose and expression, respectively. $LBS$ denotes the transformation from the canonical T-pose $T_{p}$ to the pose $\theta$, based on the pre-trained skinning weights $\mathcal{W}$ and joints $J(\beta)$. 

\noindent\textbf{3D Gaussian Splatting} \cite{kerbl20233d} explicitly represents static 3D scenes with point primitives, each primitive $G(\mathcal{X})$ could be parameterized with a 3D covariance matrix $\boldsymbol{\Sigma}$ and mean $\mu$, formulated as: $G(\mathcal{X})=e^{-\frac{1}{2}(\mathcal{X}-\mu)^T \boldsymbol{\Sigma}^{-1}(\mathcal{X}-\mu)}$. The covariance matrix could be further decomposed into scaling matrix $S$ and rotation matrix $R$ for optimization, where $\boldsymbol{\Sigma} = RSS^{T}R^{T}$. In addition, the view-dependent color is represented by coefficients of spherical harmonics. 

In this paper, we utilize diagonal vector $s \in \mathbb{R}^{3}$ and axis angle $r \in \mathbb{R}^{3}$ to represent the scaling and rotation matrix, and take RGB color $c \in \mathbb{R}^{3}$ to replace the spherical harmonic coefficients, the position of Gaussians is denoted by $\mu \in \mathbb{R}^{3}$. With the SMPL-X prior, the learning objectives of $\mu, s, r$ are defined as offsets relative to the T-pose ($\hat{\mu}, \hat{s}, \hat{r}$) template, formulated as:
$\overline{\mu}=\hat{\mu}+\mu, \overline{s}=\hat{s} \cdot s, \overline{r}=\hat{r} \cdot r$. Along with the $c$ and opacity $\alpha \in \mathbb{R}^{3}$, the canonical hman Gaussian is represented as $\mathcal{G}:$($\overline{\mu}, \overline{s},\overline{r}, c, \alpha$), which then undergo part-aware deformation via $LBS$ and output the posed human Gaussians \cite{zhang2024e3gen}. During rendering, 3D Gaussians are projected onto the pixel plane, the pixel color $C$ is calculated as: $C=\sum_{i=1}^N \alpha_i \prod_{j=1}^{i-1}\left(1-\alpha_j\right) c_i$, where $c_i$ denotes the color of the $i$-th Gaussian, and $\alpha_{i}$ is the blending weight calculated from the opacity and probability density.

\subsection{UV-Structured VAE}
\label{sec:VAE}
The VAE leverages a set of learnable tokens defined in the SMPL-X UV space to compress multi-view renderings into latent and decode the latent into 3D human Gaussians. We emphasize that an efficient and effective VAE capable of compressing 3D human Gaussians should meet the following aspects: \textbf{1)} Quickly converge under the nature of strong diversity of humans; \textbf{2)} Decoupled appearance and pose for convenient animation during subsequent generation; \textbf{3)} Sufficiently compact and low redundancy latent space, suitable for the generative model learning.

\textbf{Encoder.} VAE requires numerous steps to learn geometric structures when directly compressing multi-view renderings into 3D Gaussians \cite{henderson2024sampling}. Some works \cite{lan2024ga, yang2024atlas} introduce explicit geometric information to accelerate convergence, but the injection of explicit geometries requires the network to capture the geometric structure of different posed humans, which needs more optimization steps and may affect the convenience of subsequent pose-driven. To achieve the best of both worlds, we utilize UV-based representation \cite{zhang2024e3gen, zhuang2024idol, kwon2024generalizable} to introduce human priors and compress Gaussian attributes into learnable latent tokens, which are then decoded into Gaussian attributes and reprojected onto the canonical human Gaussians through the SMPL-X UV coordinates, and eventually obtain posed 3D human Gaussians through differentiable LBS. 

Specifically, the VAE encoder takes multi-view renderings and their corresponding Plücker \cite{sitzmann2021light} ray embeddings as the input $\mathcal{V} \in \mathbb{R}^{V \times (3+6) \times H \times W}$, where $V$ is the number of views, $H,W$ are height and width. Then $\mathcal{V}$ is fed into 3D convolutional blocks, obtaining multiview contexts $\mathbf{F}_{v} \in \mathbb{R}^{C \times V \times H^{'} \times W^{'}}$, where $H^{'}=H/8,W^{'}=W/8$. Next, we define learnable tokens $\tau \in \mathbb{R}^{N^{'} \times C}$ ( $N^{'}=H^{'}W{'}$), to accelerate the learning of appearance details (\eg, face), we select several views and project the RGB image back onto SMPL-X mesh to obtain an initial UV map \cite{tang2024intex}, which is then embedded and combined with $\tau$ as the initialization $\hat{\tau} \in \mathbb{R}^{N^{'} \times 2C}$. Subsequently, a cross-attention is employed among the $\hat{\tau}$ and flattened $\mathbf{F}_v$ to convert multi-view contexts into the latent. With the following multiple Conv-Atten dual branch blocks, we model the Gaussian latent $\tau_{L}$, the process could be expressed as:
\begin{equation}
\label{Eq.2}
\tau_{L} = \Theta(Cross(\hat{\tau}+pos, Conv(\mathcal{V}))),
\end{equation}
where $\Theta$ represents the  Conv-Atten dual branch blocks, this design captures global features and learns fine-grained local details through the inductive bias of convolution. $Cross(query,key/value)$ means the cross-attention, $pos$ is the position encoding, and $Conv$ denotes the 3D convolutional blocks. In this way, we introduce the human prior through a simple UV projection, avoiding the learning of explicit geometries and decoupling the pose and appearance. Besides, introducing learnable tokens in the UV space effectively reduces the redundancy caused by background pixels in multi-view renderings, avoiding the latent tokens being occupied to represent white pixels, which may result in the loss of representation of appearance details.

\textbf{Decoder.} The latent $\tau_{L}$ is decoded and upsampled through several 2D convolution blocks, with flowing projection heads, we obtain UV-space Gaussian attributes $\mathcal{G}_{uv}$, including the position offset $\mu \in \mathbb{R}^{N \times 3}$, scale offset $s \in \mathbb{R}^{N \times 3}$, rotation $r \in \mathbb{R}^{N \times 3}$ (equal to obtain $\overline{\mu}, \overline{s}, \overline{r}$, refer to Sec. \ref{Sec:preliminary}), opacity $\alpha \in \mathbb{R}^{N \times 1}$ and RGBs $c \in \mathbb{R}^{N \times 3}$, where $N=HW$. Then, we use the T-pose template to sample Gaussian attributes of the corresponding vertices from $\mathcal{G}_{uv}$, obtaining the canonical human Gaussians $\mathcal{G}_{c}$. Eventually, the posed human Gaussians are calculated through a forward skinning based on $LBS$. In specific, the position of each Gaussian is calculated as: $\overline{\mu}^{\prime}=\sum_{i=1}^{n_b} w_i \mathbf{B}_i \overline{\mu}$,
where $n$ is the number of joints, $w_i$ represents skinning weights, and $B_i$ denotes the transformation matrix of the $i$-th joint from the canonical pose space to the posed space. Plus, the rotation matrix is calculated as:
\begin{equation}
\mathbf{R}^{\prime}=\mathbf{T}_{1: 3,1: 3} \mathbf{R}, where \; \mathbf{T}=\sum_{i=1}^{n} w_i(\overline{\mu}) \mathbf{B}_i
\end{equation}
where $\mathbf{R}$ is the rotation matrix derived from the axis angle $r$. $T$ is the transformation matrix computed as the
weighted sum of the transformation $B_i$, $w_i({\overline{\mu}})$ corresponds to the skinning weights associated with the position $\overline{\mu}$.

\textbf{Optimization.} The VAE is optimized in an end-to-end manner, which acrosses both input views and randomly chosen views. The whole loss includes the reconstruction loss $\mathcal{L}_{\text{L1}}$, perception loss $\mathcal{L}_{\text {lpips}}$ (VGG-based), KL constraints $\mathcal{L}_{\text {KL}}$, and adversarial loss $\mathcal{L}_{\text{GAN}}$:
\begin{equation}
\mathcal{L}=\mathcal{L}_{\text {L1}}+\lambda_{\text{lpips}}\mathcal{L}_{\text {lpips }}+\lambda_{\mathrm{kl}} \mathcal{L}_{\mathrm{KL}}+\lambda_{\mathrm{GAN}} \mathcal{L}_{\mathrm{GAN}}
\end{equation}
During the first stage of the training, we use $\mathcal{L}_{\text{L1}}$, $\mathcal{L}_{\text {lpips}}$, and $\mathcal{L}_{\text {KL}}$, after number of steps, we add the adversarial loss $\mathcal{L}_{\text{GAN}}$ for the subsequent training.

\subsection{Structured Latents Generation}
\textbf{Model Design.} We employ the MM-DiT architecture \cite{kong2024hunyuanvideo, esser2024scaling, yang2024cogvideox, esser2403scaling} as the generative backbone, which treats the encoding of input conditions (\eg, text or image) and noise as a whole sequence and performs full-attention ,shown in Fig. \ref{fig:method} (b). In each Transformer block, we use Rotary Position Embedding \cite{su2024roformer} (RoPE) to enhance the model’s ability to capture both absolute and relative positional relationships. In addition, we adopt RMSNorm \cite{zhang2019root} in each block to stabilize the training. Plus, for the image-conditioned encoder, we employ a human-specialized encoder: Sapiens \cite{khirodkar2024sapiens}, which provides more fine-grained human features. For the text-conditioned generation, we take T5-XXL \cite{chung2024scaling} as the text encoder.

\textbf{Training Objective.} We employ DDPM v-prediction \cite{salimans2022progressive} to learn the transition from the conditional feature to the structured latent, the diffusion loss is formulated as:

\begin{equation}
L_{\text {gen}}(\theta):=\mathbf{E}_{t, x_0, \epsilon}\left\|\epsilon-\epsilon_\theta\left(\sqrt{\bar{\alpha}_t} x_0+\sqrt{1-\bar{\alpha}_t} \epsilon, t\right)\right\|^2
\end{equation}
where $t$ is uniformly distributed between 1 and $T$ (1000 in training), $\bar{\alpha}$ is a hyperparameter related to $t$, controlling the noise schedule, $x_0$ is the input latent, $\epsilon$ denotes the noise and $\epsilon_{\theta}$ is the network which predicts the noise. During image-to-3D training, we first use the front view as input, enabling the model to quickly learn the distribution of facial details from a large amount of data, and then use a random view as input to fine-tune the model, making it more robust to the perspective of the input image.

\section{Experiment}
\subsection{Implementation details} 
\label{Sec:imp}
For the VAE training, we render $90$ views per human Gaussian, there are $6$ input views with size $512 \times 512$ and the model is optimized by randomly choose $10$ views, the latent channel is $16$. We set $\lambda_{\text{lpips}}=1, \lambda_{\text{kl}}=1e-6$ at the first stage, and train $300k$ steps with $192$ batchsize. Then, we add $\mathcal{L}_{\text{GAN}}$ and set $\lambda_{\text{lpips}}=0.1$, using the adversarial loss dominate the decrease of perception loss and train subsequent $200k$ steps. The optimizer is AdamW \cite{loshchilov2017decoupled} with parameters $\beta_{1}=0.9,\beta_{2}=0.95$, the weight decay is $0.05$, and the learning rate is set to $2e-5$. The DiT training uses the same optimizer and parameter, with the learning rate $1e-4$, we train a total of $300k$ steps (256 batchsize). The largest model is trained with total parameters of $2$B, and take around $3500$ A$800$ GPU hours (48 GPUs). We use Classifier-Free Guidance (CFG) \cite{ho2022classifier} at inference, it is set to $3.5$ and the sampling step is $30$.

We extract $330$ samples (distinct ID) (each $20$ views) as the test set for quantitative comparison. For LGM \cite{tang2024lgm} and GHG \cite{kwon2024generalizable}, we train them using a subset of HGS-1M (around $300$k Gaussians). All quantitative experimental results are reported on this subset (including our method). For the VAE comparison, we employ three types of methods: 1) $\varphi_{1}$: directly compresses multi-view images and decodes them into multi-view Gaussians, using the video VAE \cite{yang2024cogvideox} architecture but removing its causal padding; 2) $\varphi_{2}$: this manner structures the multi-view inputs into pixel-aligned Gaussians at first, then fine-tuning an image VAE \cite{lin2025diffsplat}; 3) $\varphi_{3}$: introducing explicit human geometry as the prior, which takes posed downsampled SMPL-X points as latent points and upsamples them in the decoder to get the final human Gaussians, similar to \cite{lan2024ga}.

For the image-to-3D human Gaussian, the comparison baselines encompass several paradigms, \eg, per-optimization: SIFU \cite{zhang2024sifu}, feedforward: regression-based GHG \cite{kwon2024generalizable}, and cascade manner LGM \cite{tang2024lgm}. For the comparison of text-to-3D human Gaussian, we conduct a quantitative comparison with Diffsplat \cite{lin2025diffsplat}, please refer to the results in the Supp.

\subsection{VAE Reconstruction Results}
For a fair comparison, all comparison methods maintain the same latent size. The quantitative results of VAE comparison are shown in Tab.~\ref{table:compare} (a), our method outperforms other types across all metrics. $\varphi_{1}$ directly compresses multi-view renderings to corresponding Gaussians without introducing any priors, the limited number of latent tokens makes it hard to represent both 3D structures and appearance; $\varphi_{2}$ is a cascaded structure, if the pixel-aligned Gaussians in the first stage has low quality, it affects the final reconstruction effect. Additionally, in the pixel-aligned Gaussian, some pixels representing the rendered background are redundant, which occupy certain latent tokens, resulting in no extra tokens to represent appearance details (\eg, face). $\varphi_3$ achieves suboptimal results, but converges slowly and requires a large number of steps to grasp the structure, which affects the learning of appearance details.

\begin{table}[t]
\footnotesize
  \renewcommand{\arraystretch}{1.}
  \renewcommand{\tabcolsep}{1.pt}
\begin{subtable}[t]{0.4\linewidth}
\begin{tabular}{c|ccc}
\toprule
\textbf{Type} & \textbf{PSNR} $\uparrow$ & \textbf{LPIPS} $\downarrow$ & \textbf{SSIM} $\uparrow$ \\ \midrule
$\varphi_1$    & 24.38     & 0.067     & 0.936     \\
$\varphi_2$    & 27.74     & 0.031    & 0.941     \\
$\varphi_3$    & 28.61     & 0.028     & 0.946     \\ \midrule
\textbf{Ours}    & \textbf{29.32}     & \textbf{0.028}     & \textbf{0.953}     \\
\textbf{Ours$*$}    & \textbf{30.53}     & \textbf{0.025}     & \textbf{0.958}     \\ \bottomrule
\end{tabular}
\caption{}
\end{subtable}
\begin{subtable}[t]{0.7\linewidth}
\centering
\begin{tabular}{c|ccc}
\toprule
\textbf{Method} & \textbf{PSNR} $\uparrow$ & \textbf{LPIPS} $\downarrow$ & \textbf{SSIM} $\uparrow$ \\ \midrule
SIFU \cite{zhang2024sifu}    & 14.32     & 0.125     & 0.874     \\
LGM \cite{tang2024lgm}   & 20.87     & 0.086     & 0.925   \\
GHG \cite{kwon2024generalizable}    & 21.03     & 0.074     & 0.932     \\ \midrule
\textbf{Ours}    & \textbf{25.45}     & \textbf{0.044}     & \textbf{0.939}     \\ 
\textbf{Ours$*$}    & \textbf{26.14}     & \textbf{0.040}     & \textbf{0.943} \\
\bottomrule
\end{tabular}
\caption{}
\end{subtable}
\caption{{\textbf{(a).} Metrics of VAE comparison, where $\varphi_{1},\varphi_{2},\varphi_{3}$ represent $3$ types of methods (Sec. \ref{Sec:imp}). \textbf{(b).} Metrics of image-to-3D human Gaussians comparison, SIFU performs per-instance opimation, LGM and GHG are trained with our data. $*$ denotes the training at the whole dataset.}}
\label{table:compare}
\end{table}

\begin{figure*}[t]
	\centering
        \scriptsize
	\begin{overpic}[width=1.\linewidth]{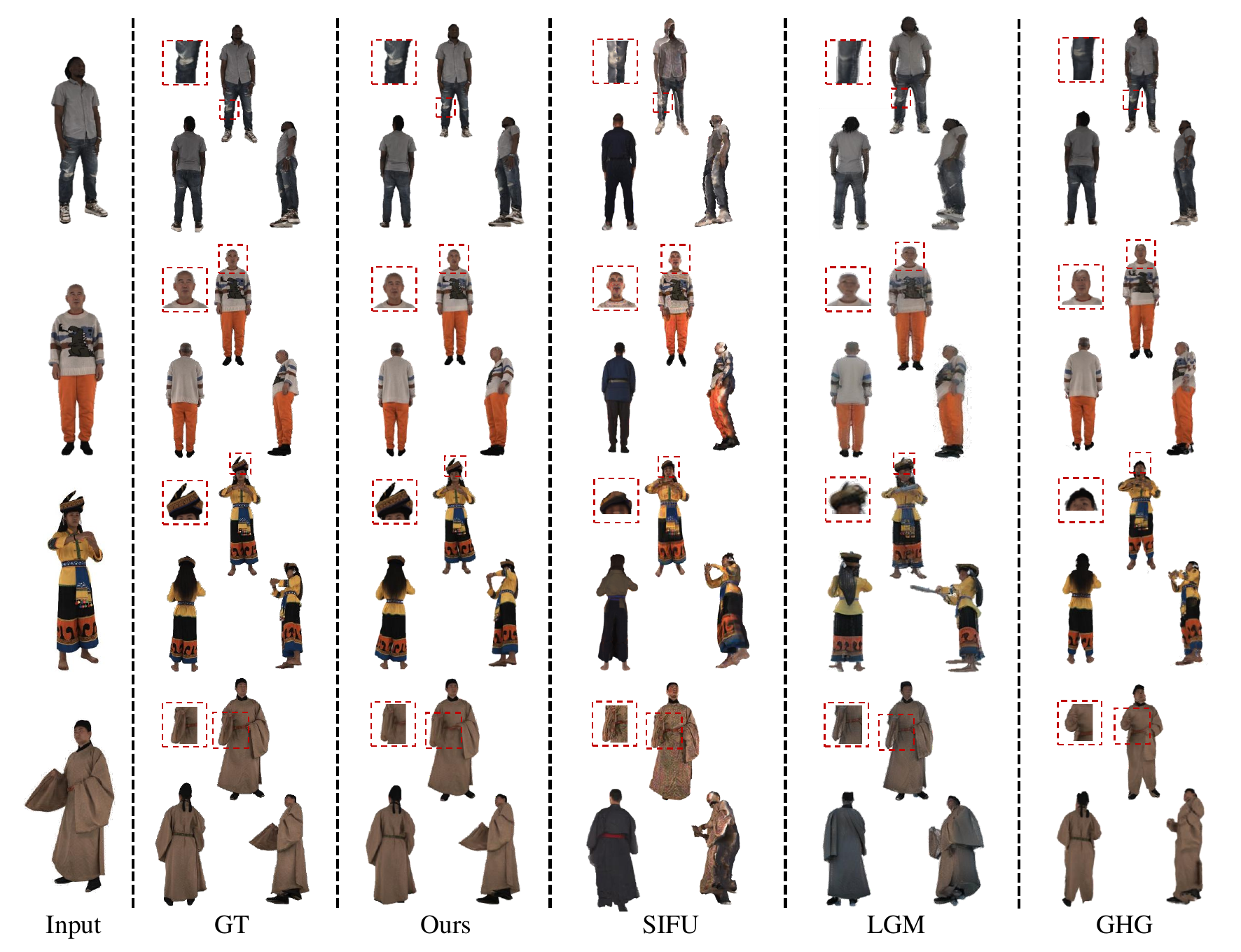}
 
	\end{overpic}
	\caption{Qualitative comparison with baselines, including inputs of different ages, races, viewpoints, and complex textures, loose cloth. Our results demonstrate more details of faces and clothes. Note: We keep the original best inference setting of the comparison method. SIFU uses one view, LGM uses four views, and GHG uses three views.}
 \label{fig:compare}
\end{figure*}

\subsection{Generation Results}
The quantitative metrics of generation are reported in Tab. \ref{table:compare} (b). Our method significantly outperforms other methods across all metrics. It shows that with our learning paradigm and method, the model can produce higher quality of 3D human Gaussians. Additionally, as shown in the Fig. \ref{fig:compare}, the quantitative results include different skin colors, genders, ages, poses, input views, complex textures, and loose clothing. The output includes horizontal, top-down, and bottom-up perspectives, which fully display the results of each method. As can be seen, our method can better express fine-grained appearance details \eg, face, diverse poses, and deformable clothing. Plus, it can be seen from the last row of Tab. \ref{table:compare} (a) and (b) that the increase in the number of data tokens can further enhance the model capabilities, reflecting that HGS-1M can support large-scale training and bring improvements in model performance.
\begin{table}[t]
\footnotesize
  \renewcommand{\arraystretch}{1.}
  \renewcommand{\tabcolsep}{1.pt}
\begin{subtable}[t]{0.4\linewidth}
\begin{tabular}{c|ccc}
\toprule
\textbf{} & \textbf{PSNR} $\uparrow$ & \textbf{LPIPS} $\downarrow$ & \textbf{SSIM} $\uparrow$ \\ \midrule
\textbf{\ding{55} $\Theta$}    & 29.04     & 0.030     & 0.950     \\
\textbf{\ding{55} ini.}    & 28.92     & 0.034    & 0.947     \\ \midrule
\textbf{LC-4}    & 25.14     & 0.042    & 0.936     \\
\textbf{LC-8}    & 28.07     & 0.037    & 0.944     \\
\textbf{LC-32}    & 29.78     & 0.026    & 0.954     \\ \midrule
\textbf{146M}    & 29.14     & 0.028     & 0.949     \\ \midrule
\textbf{Ours}    & \textbf{29.32}     & \textbf{0.028}     & \textbf{0.953}     \\ \bottomrule

\end{tabular}
\caption{}
\end{subtable}
\begin{subtable}[t]{0.7\linewidth}
\centering
\begin{tabular}{c|ccc}
\toprule
\textbf{} & \textbf{PSNR} $\uparrow$ & \textbf{LPIPS} $\downarrow$ & \textbf{SSIM} $\uparrow$ \\ \midrule
Cross.    & 24.72     & 0.052     & 0.933     \\ \midrule
CLIP   & 24.07     & 0.071     & 0.927   \\
DINOv2  & 24.95     & 0.048     & 0.933     \\ \midrule
Flow    & 25.38     & 0.046     & 0.937      \\ \midrule
200M    & 23.73     & 0.070     & 0.931      \\
1.1B    & 24.70     & 0.060     & 0.936      \\ \midrule
\textbf{Ours}    & \textbf{25.45}     & \textbf{0.044}     & \textbf{0.939} \\ 
\bottomrule
\end{tabular}
\caption{}
\end{subtable}
\caption{{\textbf{(a).} Metrics of VAE ablations, where $\Theta$ is the Conv-Atten Dual branch, ini. denotes the UV-map initialization (Sec. \ref{sec:VAE}). LC is the latent channel. \textbf{(b).} Metrics of image-to-3D human Gaussians ablations, where Cross. is the Cross-DiT, and Flow denotes training with rectified flow.}}
\label{table:ablation}
\end{table}

\subsection{Ablation Study}
\textbf{VAE.} All ablation experiments are conducted on the sub-data ($300k$), and ablation experiments of the VAE are reported in the Tab. \ref{table:ablation} (a), it could be divided into three aspects: \textbf{1) Model Design}. We validate the effectiveness of the Conv-Atten dual branch ($\Theta$ in Sec. \ref{Sec:method}) and the initialization of UV maps. For $\Theta$, we report the results of removing the convolutional branch. \textbf{2) Latent Channels}. We verify the effect of latent size on compression quality. It can be seen that the representation ability of latent space increases with the size of the latent channel. However, when the latent channel reaches $32$, although the compression quality is high, the subsequent generation model is more difficult to learn. Considering all the above, we finally use a latent size of $16$. \textbf{3) Model Scale}. Similar to the video VAE \cite{li2024wf}, we explore the impact of VAE model size on model performance, which is also shown in Tab. \ref{table:ablation} (a).

\begin{figure}
    \centering
    \small
    \begin{overpic}[width=1.0\linewidth]{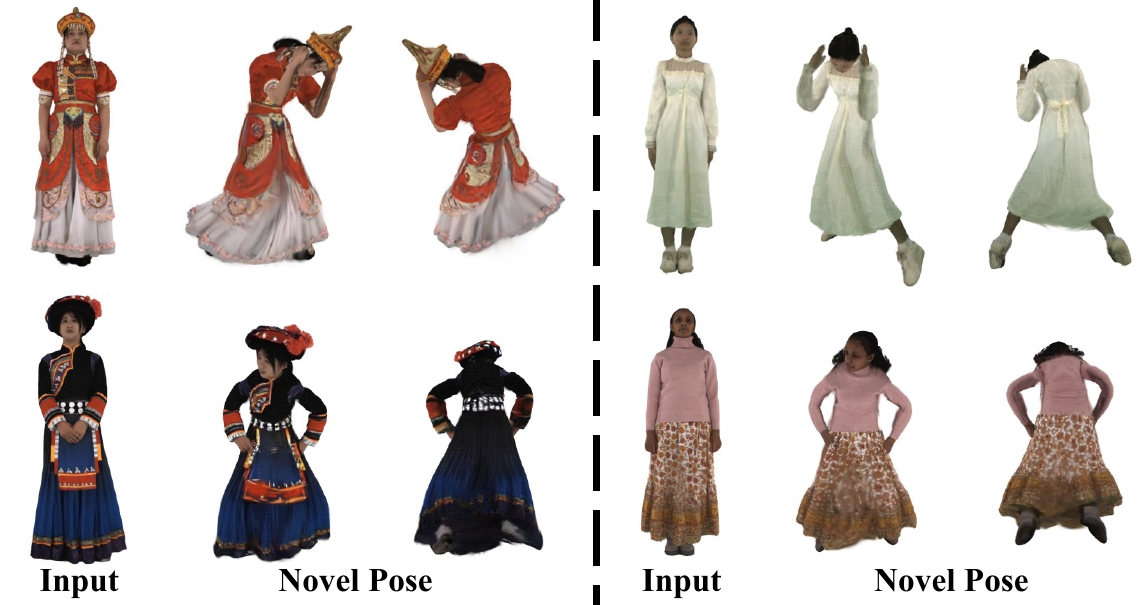}

    \end{overpic}
    \caption{The loose clothing, \eg, skirts, demonstrate nature deformation under novel poses.}
    \label{fig:deforamtion}
\end{figure}

\textbf{DiT.} The DiT ablations are reported in Tab. \ref{table:ablation} (b), encompassing $4$ aspects: \textbf{1) Model Architecture}. We conduct ablations between the Cross-DiT and MM-DiT, as can be seen from the results, for long sequence conditional feature \eg, image, MM-DiT can better model the relation between conditional features and latents. \textbf{2) Condition Encoder}. We find that in image-to-3D Human Gaussians, the condition encoder matters, we report the results when taking CLIP \cite{radford2021learning} and DINOv2 \cite{oquab2023dinov2} as the condition encoder, compared to Sapiens \cite{khirodkar2024sapiens}, they all lead to lower model performance. \textbf{3) Training Objective.} In our experiments, the difference of the final results between the DDPM v-prediction and rectified flow \cite{liu2022flow} is quite small. The difference in metrics may be due to the sampling strategy. However, we observe that DDPM converges faster in the early stage of scratch training (please refer to the Supp.). Finally, we use DDPM v-prediction for scaling up. \textbf{4) Model Scale.} Referring to the scaling-law curve of the MM-DiT architecture given in hunyuan video \cite{kong2024hunyuanvideo}, we roughly calculate the scale of the model based on the amount of data token and train models of three scales, shown in Tab. \ref{table:ablation} (b), as the model scale increases, the performance consistently improves.

\subsection{Analysis.}
\textbf{Loose clouthing deformation.} Surprisingly, the model learns the correlation between pose and loose clothing deformation from a large number of diverse poses. As can be seen from Fig. \ref{fig:deforamtion}, given an image of a person wearing loose clothing with a novel pose, we can observe that the fluffy clothing deforms naturally with the pose. This suggests that data-driven is a potential way to address this problem, and demonstrate the quality of large-scale generative models supported by large amounts of data.

\begin{figure}
    \centering
    \small
    \begin{overpic}[width=0.95\linewidth]{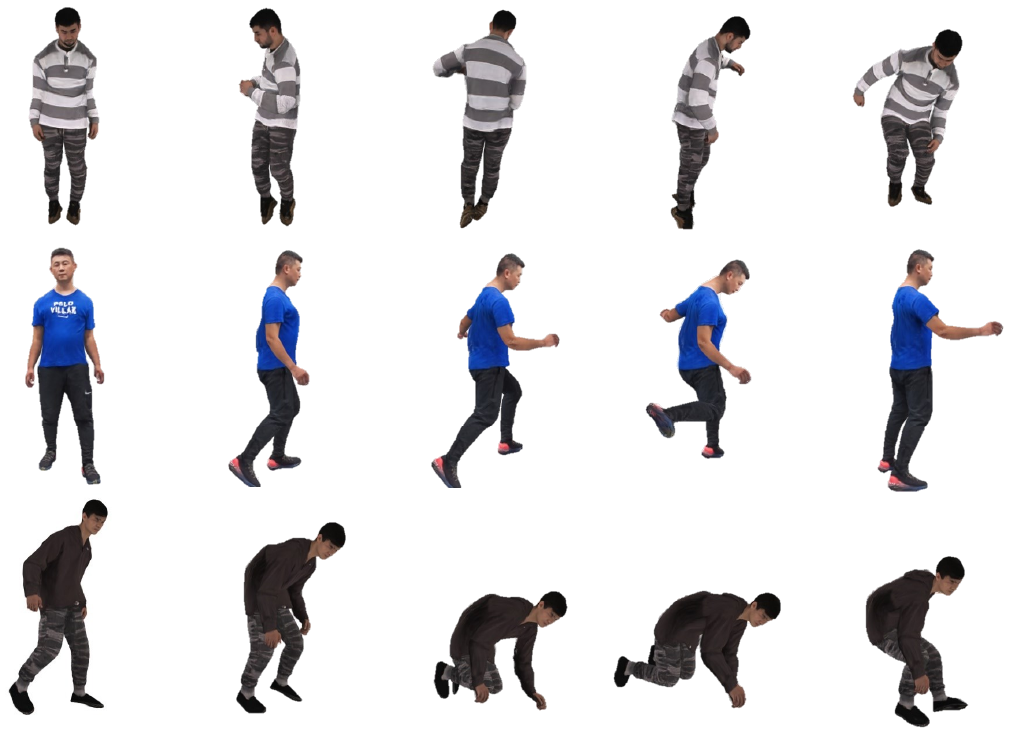}
    \end{overpic}
    \caption{The same 3D human Gaussian could be driven by different poses and maintain consistency.}
    \label{fig:motion}
\end{figure}

\textbf{Pose-Driven} Our method decouples the learning of appearance and human pose, which allows the generated human Gaussian to be driven by different poses, and the original person ID can still maintain consistency under different poses. As shown in Fig. \ref{fig:motion}, multiple people can maintain appearance consistency under multiple poses, including translation, rotation, and movement of distinct body parts.

\section{Conclusion}
In this paper, we propose the large-scale generation paradigm for 3D human digitization. To achieve this, we convert multi-view human datasets and synthetic data into a unified 3D Gaussian representation, constructing the HGS-1M dataset that contains $1$ million 3D human assets. Building upon this, we employ a UV-structured latent representation to model high-quality Gaussian latents. Subsequently, we adopt an MM-DiT architecture to learn the transfer from conditional features to structured latents, enabling the digitization of high-quality 3D humans through generation. The results demonstrate that our method can generate high-fidelity complex textures, facial details, and loose clothing from single images. This validates the feasibility and effectiveness of our proposed paradigm.

\noindent\textbf{Limitation and future work}. Currently, we train the model with a single frame, which enables us to control the human Gaussian pose, but we have not yet learned the temporal 3D Gaussians directly from the data. One of the future works could be to further generate temporal 3D Gaussians based on our HGS-1M. Additionally, multimodal control generation could be explored to advance 3D human digitization towards multi-modality controllable synthesis based on our pre-trained model.
{
    \small
    \bibliographystyle{ieeenat_fullname}
    \bibliography{main}
}
\clearpage
\section{Appendix}
\subsection{Details of HGS-1M Dataset}
\label{sec:intro}
To construct the HGS-1M dataset, we first aggregate publicly available multi-view human datasets~\cite{xiong2024mvhumannet,cheng2023dna,cai2022humman} and process them with the AnimatableGaussians framework~\cite{li2024animatablegaussians}, which optimizes dynamic human sequences into static 3D Gaussian representations. For each human sequence, we perform per-subject optimization for approximately 20 hours on an NVIDIA 4090 GPU, leveraging differentiable rendering and skeletal priors from the SMPL-X model to ensure geometric consistency across poses. The optimization process aligns each Gaussian sequence to a canonical space by extracting SMPL-X root rotation and translation parameters, then rigidly transforming all Gaussians to the origin. This step standardizes positional coordinates across diverse datasets, eliminating inconsistencies in global orientation and scale. For datasets lacking SMPL-X annotations (e.g., synthetic assets), we fit SMPL-X parameters using EasyMoCap~\cite{easymocap} before alignment.

After alignment, we render each human Gaussian from 90 viewpoints to maximize supervision coverage. The camera setup includes 30 horizontal views (azimuth angles spaced at 12° intervals), 30 upward views (elevation +15° to +90°, azimuth spaced at 30° intervals), and 30 downward views (elevation -15° to -90°), ensuring dense angular sampling for full 360° reconstruction. For small-scale datasets like Thuman2.1 \cite{yu2021function4d} and 2K2K \cite{han2023high}, we apply identical rendering protocols to maintain consistency. To further enhance diversity, we integrate 100k synthetic human assets generated via parametric body models (e.g., SMPL-X) and procedurally augmented with varied textures, clothing meshes, and lighting conditions. These synthetic assets are converted into Gaussians using the same optimization pipeline, with their root transformations reset to the origin. The final dataset combines optimized real-world sequences, rendered multi-view data, and synthetic samples, totaling 1 million human 3D Gaussians.

\subsection{Method Details}
\textbf{VAE}. The encoder of VAE that accepts multi-view input consists of 4 3D-convolutional blocks, which downsamples the $H$ and $W$ of the original image by $8$ times. The view dimension is not downsampled, and the input image size is $512\times512$. For learnable tokens, the initial width and height are the same as the size after the VAE 3D convolution encoder downsamples. In addition, for the initialized UV map, we selected 16 viewpoints to project back the RGB value to the mesh and extract the output UV map value. After that, this UV map is encoded through a $1 \times 1$ 2D convolutional block into the feature dimension of the learnable token and concatenated with it as the final initialized token. After cross attention, we use $6$ Conv-Attn dual branch blocks to model the latent, and the final latent size is $64 \times 64 \times 16$. The VAE decoder includes $4$ 2D convolution blocks, and finally upsamples the UV map with Gaussian attributes to $512 \times 512$ for sampling. After this, multiple decode heads corresponding to various Gaussian attributes are employed to obtain the final Gaussian.

\begin{figure}[t]
	\centering
        \scriptsize
	\begin{overpic}[width=1.\linewidth]{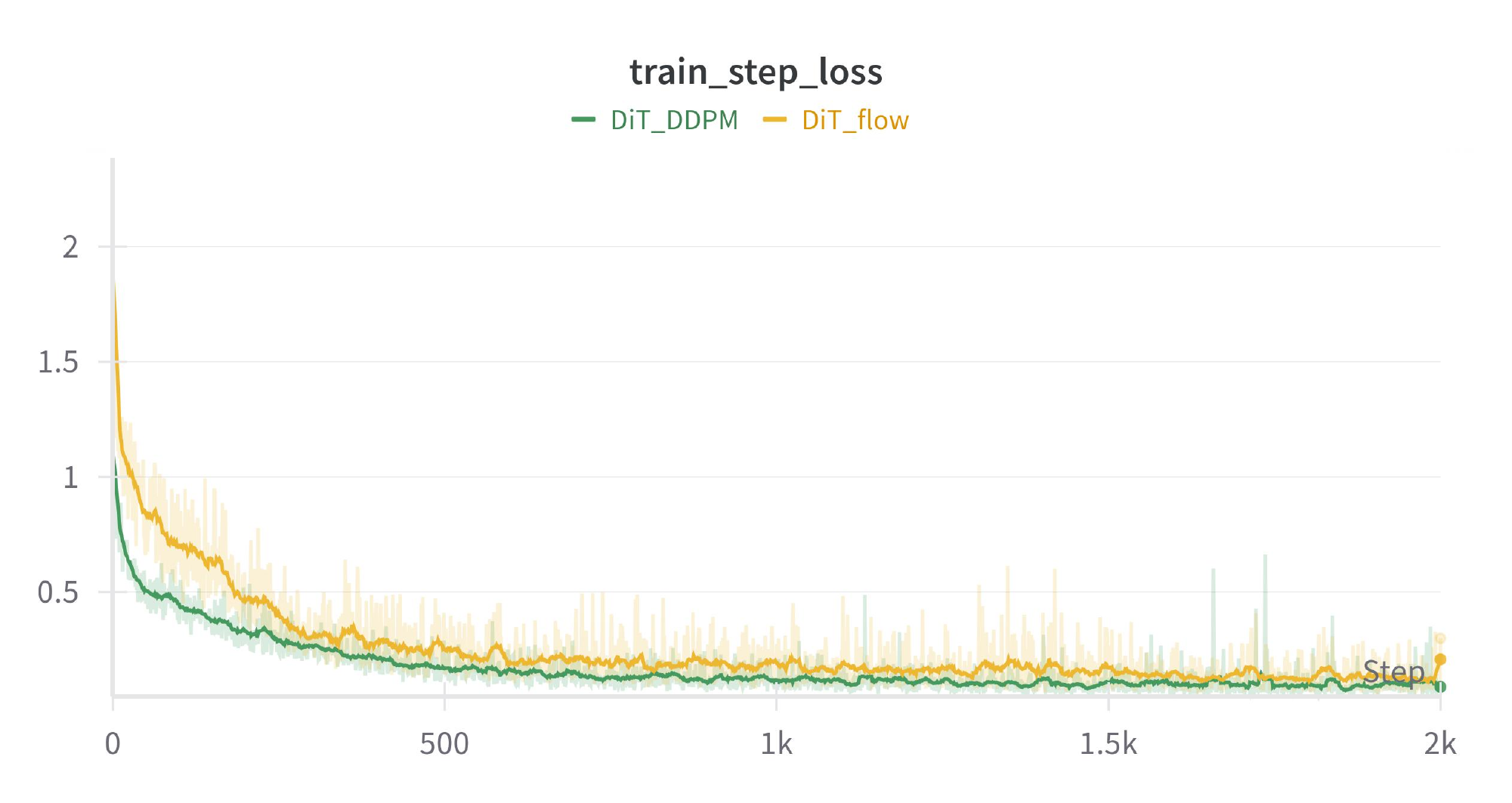}
 
	\end{overpic}
	\caption{Changes in DDPM v-prediction and Rectified flow training loss curves in the early stages of training}
 \label{fig:train}
\end{figure}

\textbf{MM-DiT} We use 2D rotation position encoding (RoPE) and RMS-norm in the DiT architecture. According to our observation, it is necessary to add RMS-Norm to the training from scratch. After that, the model is more robust to the learning rate and can converge normally under multiple learning rates. For the final 2B model, a total of $30$ MM-DiT blocks are included, with $32$ attention heads in each block, for a total of $64$ heads. For the DIT training objective, in our experiments, the difference between using DDPM and flow matching is not obvious. For flow matching, we use the same noise addition method and sampling shift strategy as SD3 \cite{esser2024scaling}, and finally do not observe a significant performance improvement over DDPM. We observe that DDPM converges faster in the early stage when training from scratch, as shown in the Fig. \ref{fig:train}.

\begin{figure*}[t]
	\centering
        \scriptsize
	\begin{overpic}[width=1.\linewidth]{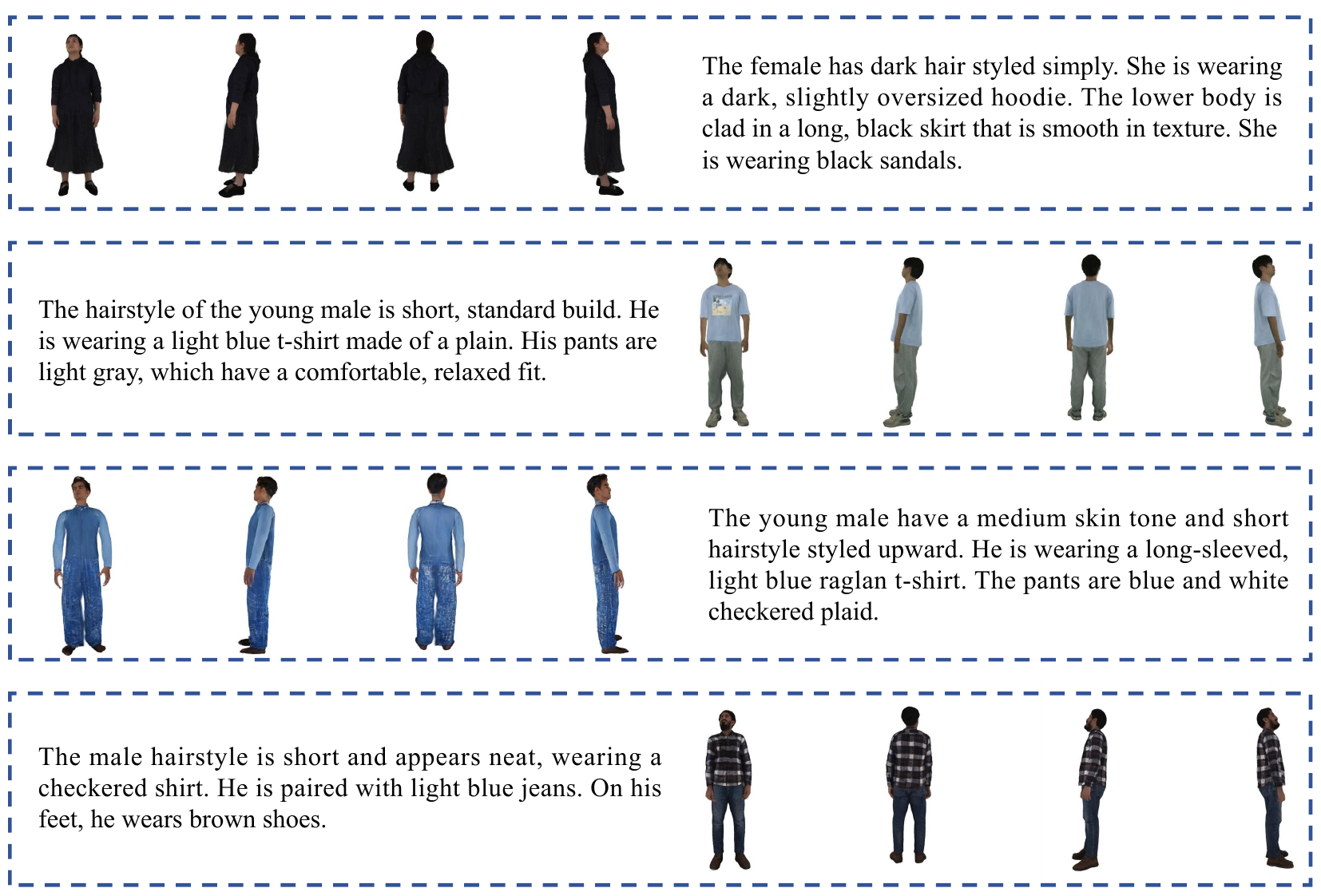}
 
	\end{overpic}
	\caption{Results of Text-to-3D human Gaussians.}
 \label{fig:supp_text}
\end{figure*}

\textbf{Baselines}. \textbf{1) GHG} \cite{kwon2024generalizable}: The official checkpoint released by GHG \cite{kwon2024generalizable} is trained on the THuman2.0 \cite{yu2021function4d} dataset, which contains approximately 500 3D human subjects. To ensure a fair comparison, we fine-tune GHG on part of the HGS-1M dataset. For each subject, we render nine views evenly distributed across azimuth and altitude. For the inpainting network, we directly use the official checkpoint. During training, following GHG, we use three fixed horizontal views as input and apply multi-view supervision on three randomly sampled views with evenly distributed azimuth. During inference, we use the same three input views to generate novel viewpoints.
\textbf{2) LGM} \cite{tang2024lgm}, if we remove the step of generating multi-view images from a single image in front of LGM, LGM is a process of outputting Gaussian images from multiple perspectives. Given this, we train its second stage and directly give $4$ GT perspectives during inference, removing the multi-view generation step. In theory, LGM can output the best results. \textbf{3)} SIFU \cite{zhang2024sifu} is a per-subject optimization method for single-image 3D human reconstruction. For each subject, we use the front view as input to generate a textured 3D human model. \textbf{DiffSplat} \cite{lin2025diffsplat}, We use Diffsplat to compare the text-to-3D Gaussian. We use its original VAE and fine-tune it with our text and 3D data pairs. Finally, we calculated the CLIP score \cite{radford2021learning} indicator on $100$ test samples. The result of our model is $25.89$, while the result of DiffSplat is $24.62$. We also provide some visual results of the text-to-3D, generated by our methods, please refer to Fig. \ref{fig:supp_text}.

\begin{figure*}[t]
	\centering
        \scriptsize
	\begin{overpic}[width=1.\linewidth]{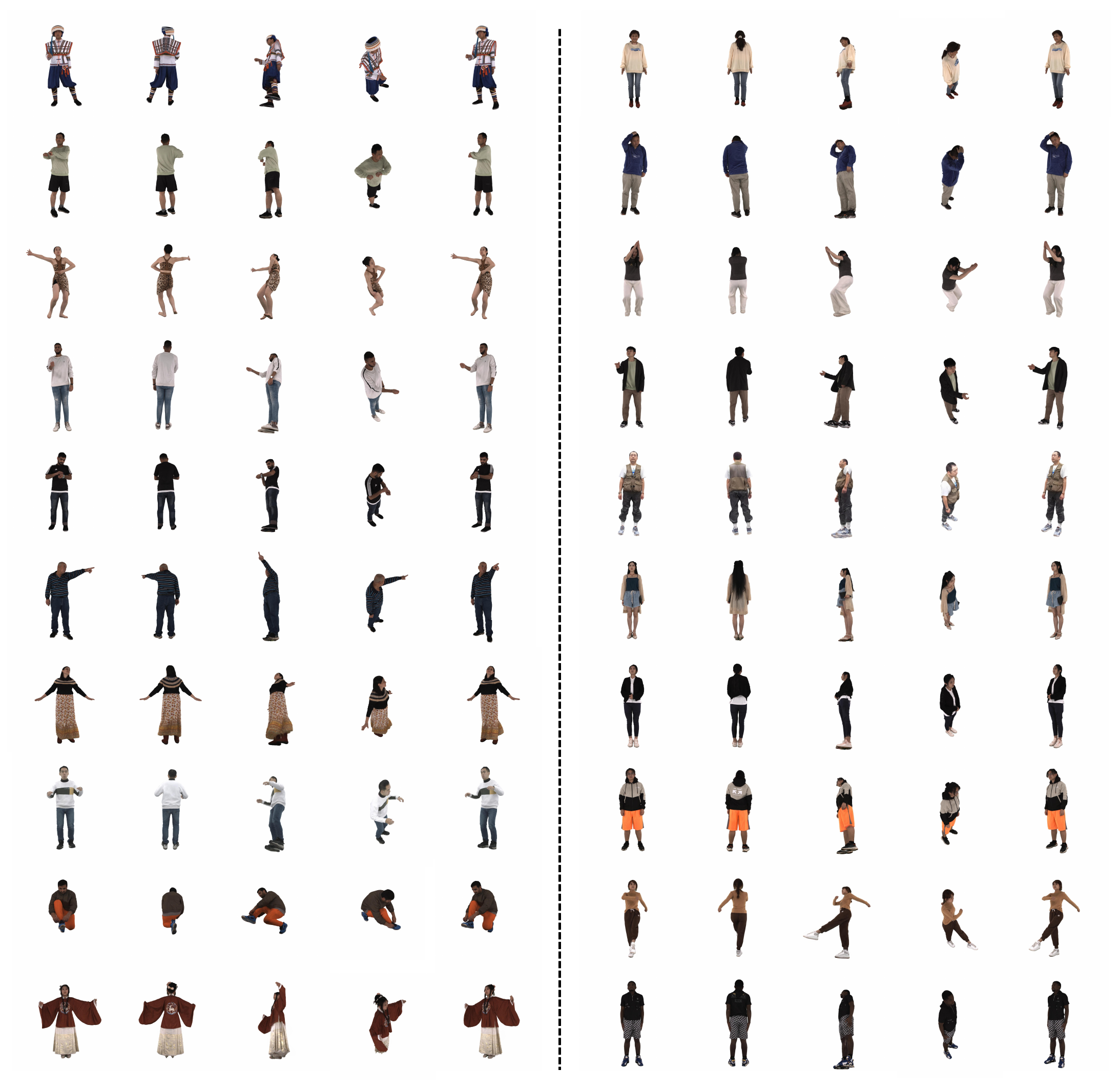}
 
	\end{overpic}
	\caption{Visualization of sampled cases from our HGS-1M Dataset.}
 \label{fig:supp_data}
\end{figure*}

\begin{figure*}[t]
	\centering
        \scriptsize
	\begin{overpic}[width=1.\linewidth]{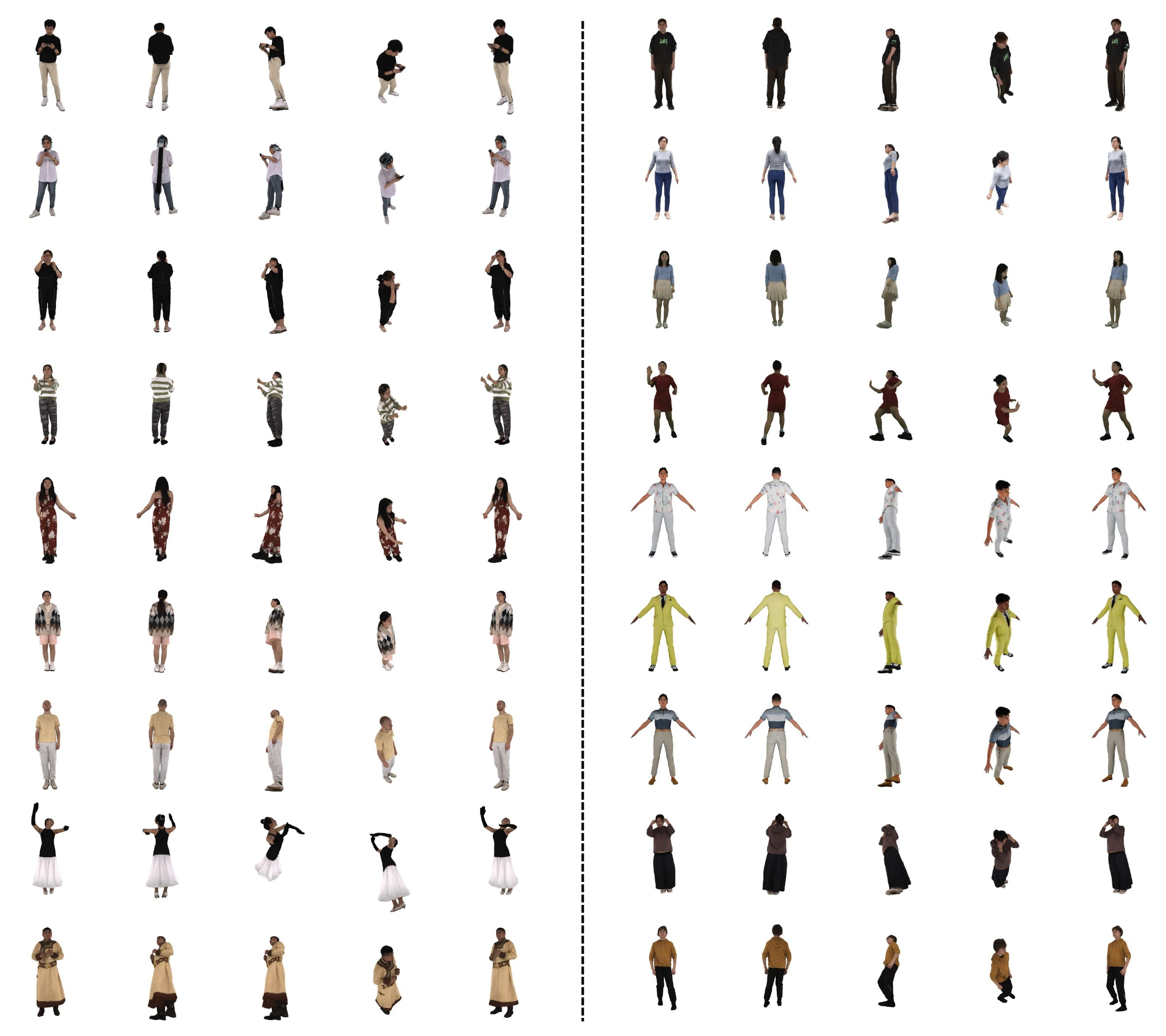}
 
	\end{overpic}
	\caption{Results of single image-to-3D human Gaussians.}
 \label{fig:supp_img}
\end{figure*}

\end{document}